\documentclass{article}

\usepackage{microtype}
\usepackage{graphicx}
\usepackage{amsmath}
\usepackage{amssymb}
\usepackage{booktabs} 
\usepackage{subcaption}

\usepackage{hyperref}

\usepackage[accepted]{icml2019}

\icmltitlerunning{`In-Between' Uncertainty in Bayesian Neural Networks}

\begin{document}

\twocolumn[
\icmltitle{`In-Between' Uncertainty in Bayesian Neural Networks}




\begin{icmlauthorlist}
\icmlauthor{Andrew Y. K. Foong}{cam}
\icmlauthor{Yingzhen Li}{microsoft}
\icmlauthor{Jos\'e Miguel Hern\'andez-Lobato}{cam,microsoft,turing}
\icmlauthor{Richard E. Turner}{cam,microsoft}
\end{icmlauthorlist}

\icmlaffiliation{cam}{Department of Engineering, University of Cambridge, Cambridge, United Kingdom}
\icmlaffiliation{microsoft}{Microsoft Research Cambridge, United Kingdom}
\icmlaffiliation{turing}{Alan Turing Institute, London, United Kingdom}

\icmlcorrespondingauthor{Andrew Foong}{ykf21@cam.ac.uk}

\icmlkeywords{Machine Learning, ICML, Bayesian Neural Networks, Regression, Approximate Inference, Uncertainty}

\vskip 0.3in
]



\printAffiliationsAndNotice{}  

\begin{abstract}
We describe a limitation in the expressiveness of the predictive uncertainty estimate given by mean-field variational inference (MFVI), a popular approximate inference method for Bayesian neural networks. In particular, MFVI fails to give calibrated uncertainty estimates in between separated regions of observations. This can lead to catastrophically overconfident predictions when testing on out-of-distribution data. Avoiding such overconfidence is critical for active learning, Bayesian optimisation and out-of-distribution robustness. We instead find that a classical technique, the linearised Laplace approximation, can handle `in-between' uncertainty much better for small network architectures. 
\end{abstract}

\section{Introduction}

Neural networks have been shown to be extremely successful for supervised learning. However, they are known to underestimate their uncertainty when trained by maximum likelihood or maximum a posteriori (MAP) methods. A neural network that returns reliable uncertainty estimates whilst maintaining the computational and statistical efficiency of standard networks would have numerous applications in active learning, reinforcement learning and critical decision-making tasks \cite{gal2017deep,chua2018deep,gal2016uncertainty}. 

A variety of techniques have been proposed to obtain uncertainty estimates for neural networks in computationally efficient ways \cite{mackay1992practical,hinton1993keeping,barber1998ensemble,hernandez2015probabilistic,gal2016dropout,lakshminarayanan2017simple}. Among these, mean-field variational inference (MFVI) is a widely used approximate inference method that gives state-of-the-art performance in non-linear regression. On the commonly used UCI regression benchmark, MFVI with the reparameterisation trick \cite{blundell2015weight,kingma2015variational} often outperforms Stochastic Gradient Langevin Dynamics, Probabilisitic Back-Propagation and ensemble methods \cite{bui2016deep,tomczakneural}, and is competitive with Monte Carlo Dropout.\footnote{See \cite{mukhoti2018importance} for a recent strong baseline for Monte Carlo Dropout on the UCI regression datasets.}

Performance on the UCI datasets is usually measured by held out log-likelihood. This represents both accuracy and uncertainty quantification, since it penalises methods that are overconfident in addition to being inaccurate. It is therefore perhaps surprising that MFVI performs poorly on sequential decision making tasks that require good uncertainty quantification, such as contextual bandits \cite{riquelme2018deep}. To perform well, a method must `know what it knows, and what it doesn't know': it should have high confidence near clusters of observations, and low confidence elsewhere. More specifically, a well-calibrated network should predict with high uncertainty far from data, as well as \emph{in regions between separated clusters of observations}. However, the current UCI benchmark is not suitable for evaluating `in-between' uncertainty, as the test set is obtained by uniformly subsampling the full dataset. We therefore design another UCI benchmark to test for in-between uncertainty, by taking the `middle region' of the full dataset as the test set. We find that although MFVI performs well on the standard UCI benchmark, it can fail catastrophically on the in-between version, showing the detrimental effect the mean-field approximation has on in-between uncertainty. In contrast, a classical technique, the linearised Laplace approximation \cite{mackay1992practical}, performs well on both.

\section{Methods}

This paper focuses on two approximate inference techniques for Bayesian Neural Networks (BNNs): Variational Inference (VI) and the Laplace approximation. We consider networks whose output $f_{\theta}(\mathbf{x})$ given input $\mathbf{x}$ and parameters $\theta$ is interpreted as the mean of a Gaussian distribution with homoscedastic output noise variance $\sigma_o^2$. We place a diagonal Gaussian prior over $\theta$, here written as a column vector of all weights and biases in the network.

\textbf{Variational Inference}. Let the posterior distribution over $\theta$ given a dataset $\mathcal{D} = \{ (\mathbf{x}_n, y_n) \}_{n=1}^N$ be $p(\theta| \mathcal{D})$. VI approximates this posterior with a simpler distribution $q_{\phi}(\theta)$. The parameters $\phi$ are learned by optimising a simple Monte Carlo estimate of the Evidence Lower Bound (ELBO):
\begin{align*}
    \mathcal{L}(\phi) &= \sum_{n=1}^N \mathbb{E}_{q_{\phi}}[ \log p(y_n|\theta, \mathbf{x}_n) ] - \mathrm{KL}(q_{\phi}(\theta)||p(\theta))\\
    &\approx \frac{1}{M} \sum_{m=1}^M \sum_{n=1}^N \log p(y_n|\theta_m, \mathbf{x}_n) - \mathrm{KL}(q_{\phi}(\theta)||p(\theta)),
\end{align*}
where $\theta_m$ are sampled from $q_{\phi}(\theta)$. Optimising the ELBO minimises the KL-divergence between $q_{\phi}(\theta)$ and the true posterior. Once $\phi$ is learned, we can make predictions by Monte Carlo sampling from $q_{\phi}(\theta)$:
\begin{align}
    p(y^*|\mathbf{x}^*, \mathcal{D}) &= \mathbb{E}_{p(\theta|\mathcal{D})}[p(y^*|\mathbf{x}^*, \theta)] \nonumber\\
    &\approx \mathbb{E}_{q_{\phi}(\theta)}[p(y^*|\mathbf{x}^*, \theta)]\nonumber\\
    &\approx \frac{1}{M} \sum_{m=1}^M p(y^*|\mathbf{x}^*, \theta_m). \label{MC_predictive}
\end{align}
A common and scalable choice for the form of $q_{\phi}(\theta)$ is the mean-field Gaussian approximation (MFVI), which is a fully factorised Gaussian distribution. Another choice is to let $q_{\phi}(\theta)$ be a full covariance Gaussian (FCVI). This is more flexible, but the number of variational parameters $\phi$ is now quadratic in the number of parameters in the network. 

\textbf{Laplace Approximation}. The Laplace approximation \cite{denker1991transforming,mackay1992practical} finds a mode $\theta_{\mathrm{MAP}}$ of the posterior, and sets the approximate posterior to $q(\theta) = \mathcal{N}(\theta; \mu, \Sigma)$ with $\mu = \theta_{\mathrm{MAP}}$. $\Sigma$ is set such that the curvature of $\log p(\theta|\mathcal{D})$ matches the curvature of the logarithm of the Gaussian approximation at $\theta_{\mathrm{MAP}}$, that is:
\begin{align*}
    \Sigma = -\bigg[ \nabla_{\theta} \nabla_{\theta} \log p(\theta|\mathcal{D}) \big\rvert_{\theta = \theta_{\mathrm{MAP}}} \bigg]^{-1}.
\end{align*}
In words, $\Sigma$ is the negative inverse Hessian evaluated at the MAP solution. In practice we use the Gauss-Newton matrix, which is guaranteed to be positive semi-definite, and can be evaluated using only first derivatives: 
\begin{align*}
    \Sigma = -\bigg[ \frac{1}{\sigma_o^2} \sum_{n=1}^N g(\mathbf{x}_n) g(\mathbf{x}_n)^\mathsf{T} + \mathrm{diag}(p) \bigg]^{-1}.
\end{align*}
Here $g(\mathbf{x}_n) = \nabla_{\theta} f_{\theta}(\mathbf{x}_n) \big\rvert_{\theta = \theta_{\mathrm{MAP}}}$ and $p$ is a vector whose $i$th element is $1/\sigma^2_i$, where $\sigma^2_i$ is the prior variance of $\theta_i$.

Once $\theta_{\mathrm{MAP}}$ and $\Sigma$ are obtained, there are two different ways to make predictions. The first is to Monte Carlo sample from the approximate posterior as in equation (\ref{MC_predictive}). We refer to this method as Sampled Laplace (SL). Unfortunately, the Laplace approximation is known to cause severe underfitting \cite{lawrence2001variational}. An alternative procedure which empirically alleviates this is to linearise the output of the network about $\theta_{\mathrm{MAP}}$. This leads to a linear Gaussian model that can be solved exactly for the predictive distribution:
\begin{align*}
    p(y^*|\mathbf{x}^*, \mathcal{D}) \approx \mathcal{N}(y^*; f_{\theta_{\mathrm{MAP}}}(\mathbf{x}^*), \sigma_o^2 + g(\mathbf{x}^*)^{\mathsf{T}}\Sigma \, g(\mathbf{x}^*));
\end{align*}
see Appendix \ref{laplace_linearised} for details. We refer to this method as Linearised Laplace (LL).

Finding $\theta_{\mathrm{MAP}}$ is identical to standard neural network training. Once at a mode, calculating the Gauss-Newton matrix requires one backward pass for each element of the dataset, which has a cost that scales linearly in the number of observations. Lastly the Laplace approximation requires inverting this matrix which has cubic cost in the number of parameters. This is still tractable for the smaller networks typically considered for regression on UCI datasets. Recent work has applied Kronecker-Factored Approximate Curvature (K-FAC) to obtain a scalable method; however they used sampling instead of linearisation and had to take steps to mitigate the underfitting problem inherent in the Laplace approximation \cite{ritter2018scalable}.

\section{Experiments}
\label{experiment_section}
To test for in-between uncertainty, we compare these methods on two tasks. The first is a synthetic 1D regression dataset formed by adding Gaussian noise to a sine wave and observing two separated clusters of input points. The second are the UCI regression datasets.

\begin{figure*}
\vskip 0.2in
\begin{center}
\centerline{\includegraphics[width=2\columnwidth]{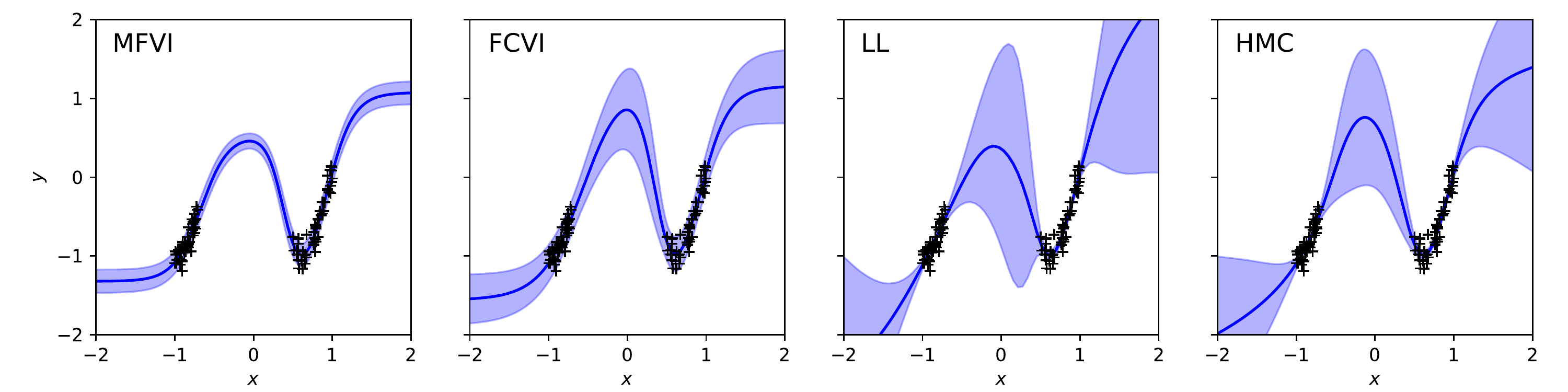}}
\caption{Mean and two standard deviation bars of the predictive distribution for $f_{\theta}(x)$ (without output noise). }
\label{tanh_comparison}
\end{center}
\vskip -0.2in
\end{figure*}

\textbf{Synthetic 1D Dataset}. We plot results for four inference methods on the 1D dataset: MFVI, FCVI, LL and Hamiltonian Monte Carlo (HMC), in Figure \ref{tanh_comparison}.\footnote{SL is not shown as the fit is so poor that the error bars completely fill the figure. See \cite{lawrence2001variational} pp. 88 - 91.} We use a single hidden layer network with tanh non-linearities\footnote{ReLUs caused problems with LL, see Appendix \ref{appendix_extra}.} and 50 hidden units. Diagonal Gaussian priors are used for all networks, and the observation noise $\sigma_o$ is fixed to $0.1$, the true value. Details and additional results are in Appendix \ref{appendix_extra}. We see that MFVI fails to represent in-between uncertainty: its error bars are of similar magnitude in the data region and the in-between region. FCVI has larger uncertainty in the middle, but is slightly underconfident in the data region. LL and HMC show high confidence in the data region and increased uncertainty in between, showing that MFVI's failure is rooted in approximate inference, not the model class. 

There are several reasons for MFVI's overconfidence. First, we show in Appendix \ref{convex_variance} that a single hidden layer BNN with ReLU activations and with deterministic input weights and mean-field (possibly non-Gaussian) output weights must have an output variance that is \emph{convex} as a function of its input. Such a BNN is incapable of expressing increased uncertainty between regions of low uncertainty. This would not be the case if the output weights had an unrestricted distribution. Although this insight does not immediately apply to BNNs with tanh activations and mean-field input weights, it shows that the mean field assumption can in some cases severely restrict the complexity of uncertainty estimates a BNN can express in function space.

Second, MFVI fails to express increased in-between uncertainty because fitting data in the outer region whilst having increased uncertainty in-between requires strong \emph{dependencies} in the approximate posterior. This is because in a mean-field distribution, any parameter uncertainty used to express increased in-between uncertainty leads to uncontrolled variations in the fit in the data region. The only way to have a good fit \emph{and} increased in-between uncertainty is to have variations in one parameter compensated for by variations in others, such that the resulting function still passes through the data points. This is explained in detail in Appendix \ref{toy_explanation} via a synthetic example.

\begin{figure*}
\vskip 0.2in
\begin{center}
\centerline{\includegraphics[width=2\columnwidth]{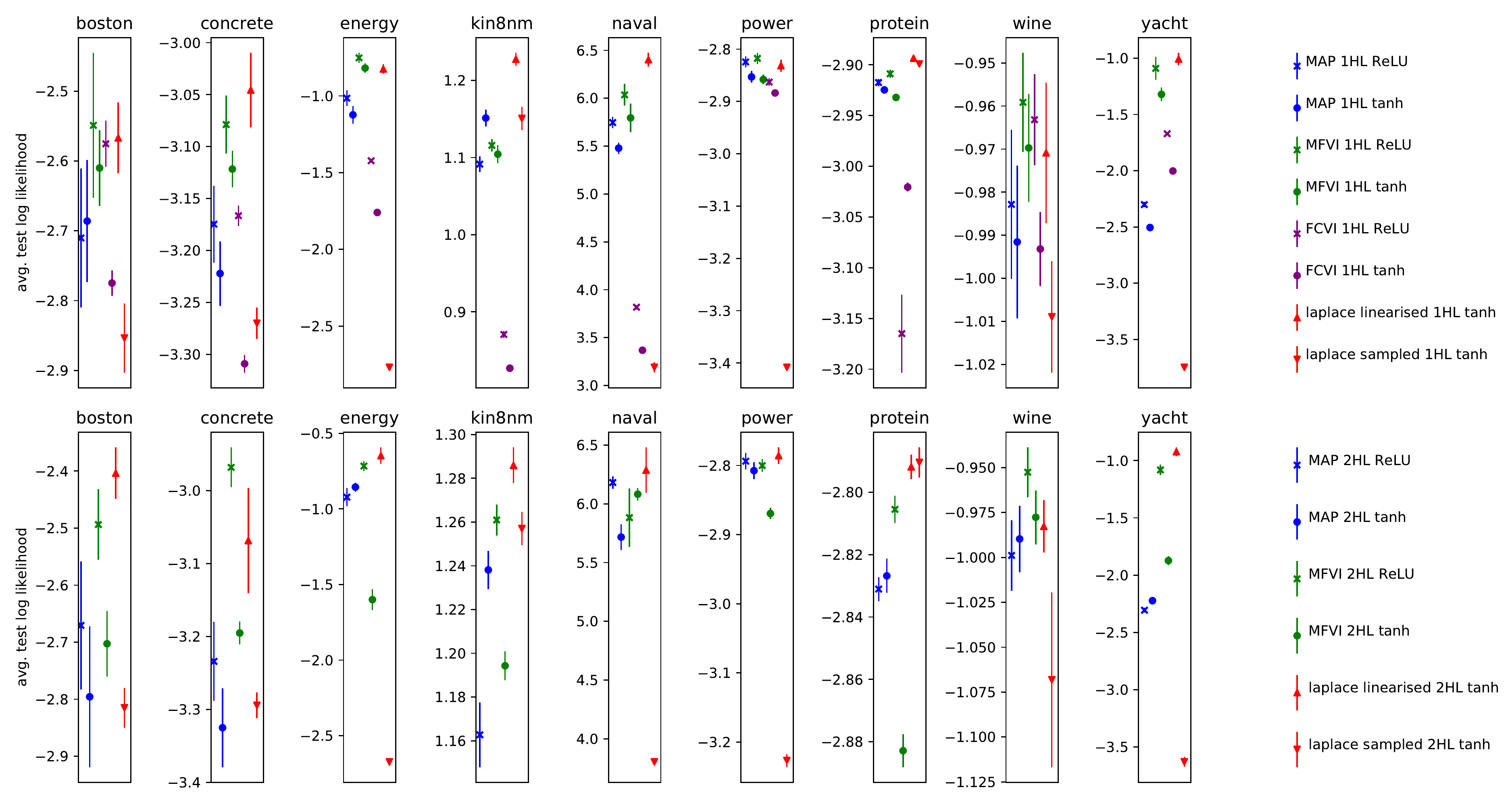}}
\caption{Average test log-likelihoods on the standard splits for BNNs with one hidden layer (top) and two hidden layers (bottom). There are 50 hidden units in each layer.}
\label{error_bars_standard}
\end{center}
\vskip -0.2in
\end{figure*}

\begin{figure*}
\vskip 0.2in
\begin{center}
\centerline{\includegraphics[width=2\columnwidth]{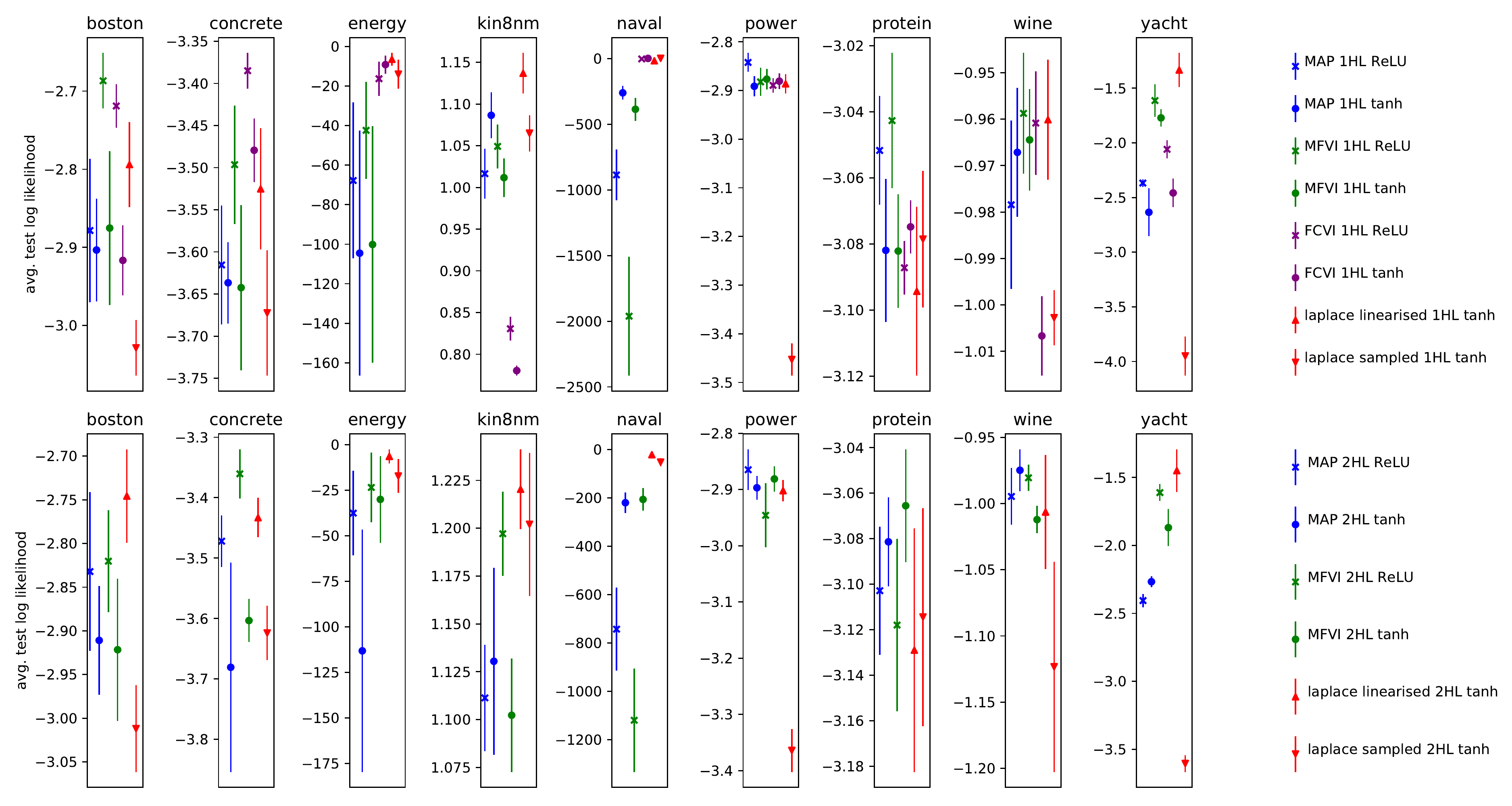}}
\caption{Average test log-likelihoods on the gap splits for BNNs with one hidden layer (top) and two hidden layers (bottom). Note the scale on energy and naval, where MAP and MFVI fail catastrophically. There are 50 hidden units in each layer.}
\label{error_bars_gap}
\end{center}
\vskip -0.2in
\end{figure*}

\textbf{UCI Regression Datasets}. We now investigate the uncertainty quality of BNNs in real-world datasets. The UCI datasets are usually split into training and test sets by subsampling the dataset uniformly. In the first experiment, we use the standard splits also used in \cite{hernandez2015probabilistic,bui2016deep,mukhoti2018importance}. In our second experiment, we create custom splits to test for in-between uncertainty. For each of the $D$ input dimensions of $\mathbf{x}_n \in \mathbb{R}^D$, we sort the datapoints in increasing order in that dimension. We then remove the middle $1/3$ of these datapoints for use as a test set. The outer $2/3$ are the training set. $10\%$ of the training set is used as a validation set. Thus a dataset with $D$ inputs has $D$ splits. We refer to these as the `gap splits'. A satisfactory method would achieve good results on the standard splits (showing an ability to fit the data well) and avoid catastrophically poor results on the gap splits (showing increased in-between uncertainty). The results are shown in Figures \ref{error_bars_standard} and \ref{error_bars_gap}.\footnote{FCVI was only run on one hidden layer due to its long training times, and only tanh was used for Laplace as ReLUs caused problems with linearisation - see Figure \ref{relu_comparison}. Full UCI results and experimental details in Appendix \ref{appendix_extra}.}

For the standard splits MFVI and LL perform best. FCVI does poorly, likely due to optimisation difficulties. LL does much better than SL, which is surprising given linearisation adds another approximation. However, it appears to redeem the poor Gaussian approximation \cite{lawrence2001variational}. For the gap splits, MAP is competitive with Bayesian methods on power, protein and wine. However, on energy and naval MAP fails catastrophically, doing dozens or hundreds of nats worse than LL. The test sets of energy and naval thus show very different behaviour from their training sets, and good in-between uncertainty is required to prevent overconfident extrapolations. In this situation we would expect Bayesian methods to outperform MAP. However MAP and MFVI perform similarly poorly on energy and naval, showing that MFVI is overconfident. The only method that performs well on the standard splits and avoids any catastrophic results on the gap splits is LL.

\section{Conclusions}

We have shown that MFVI fails to provide calibrated in-between uncertainty, and that the standard UCI splits fail to adequately test for it. However, the decades-old LL approximation performs far better in this regard. Although recent advances in variational inference have allowed BNNs to scale to larger architectures than ever before, in terms of uncertainty \emph{quality} the mean-field approximation loses crucial expressiveness compared to the less scalable LL approximation. It is therefore key for the field of approximate inference to consider how the approximation of posteriors in parameter space affects the expressiveness of uncertainties in \emph{function space}. Future work will investigate the conditions an approximate posterior must satisfy to reliably capture in-between uncertainty. It would also be natural to see if combining K-FAC Laplace \cite{ritter2018scalable} with linearisation leads to improved results.

\section*{Acknowledgements}


We thank David R. Burt, Sebastian W. Ober and Ross Clarke for helpful discussions. AYKF gratefully acknowledges the Trinity Hall Research
Studentship and the George and Lilian Schiff Foundation for funding his studies.


\bibliography{example_paper}
\bibliographystyle{icml2019}

\appendix
\section{Linearised Laplace Approximation}
\label{laplace_linearised}
To obtain the linearised Laplace approximation, we linearise the output of the network about $\theta_{\mathrm{MAP}}$:
\begin{align}
    f_{\theta}(\mathbf{x}) \approx f_{\theta_{\mathrm{MAP}}}(\mathbf{x}) + g(\mathbf{x})^{\mathsf{T}}(\theta - \theta_{\mathrm{MAP}}). \label{linearisation}
\end{align}
We now have the following approximating distributions:
\begin{align*}
    p(\theta|\mathcal{D}) &\approx  \mathcal{N}(\theta; \theta_{\mathrm{MAP}}, \Sigma),\\
    p(y^*|\theta, \mathbf{x}^*) &\approx \mathcal{N}(y^*; f_{\theta_{\mathrm{MAP}}}(\mathbf{x}) + g(\mathbf{x})^{\mathsf{T}}(\theta - \theta_{\mathrm{MAP}}), \sigma_o^2).
\end{align*}
Since this is now a linear-Gaussian model, we can use standard formulas to obtain:
\begin{align*}
    p(y^*|\mathbf{x}^*, \mathcal{D}) \approx \mathcal{N}(y^*; f_{\theta_{\mathrm{MAP}}}(\mathbf{x}^*), \sigma_o^2 + g(\mathbf{x}^*)^{\mathsf{T}}\Sigma \, g(\mathbf{x}^*)).
\end{align*}

\section{Convex Variance Result}
\label{convex_variance}
 Consider a single hidden layer BNN with input $\mathbf{x} \in \mathbb{R}^D$ and output $\mathbf{y} \in \mathbb{R}^K$ with a mean field distribution over the output weights and biases $(\mathbf{W}, \mathbf{b})$ but a point estimate for the input weights and biases $(\mathbf{U}, \mathbf{v})$. In detail:
\begin{align*}
    y_k (\mathbf{x}) &= \sum_i W_{ki}\phi(a_i) + b_k,\\
    a_i &= \sum_j U_{ij}x_j + v_i.
\end{align*}
We assume a fully factorised approximating distribution for the output weights such that: 
\begin{align*}
    q(\mathbf{W}, \mathbf{b}) = q(\mathbf{b})\prod_{k,i}q_{ki}(W_{ki}).
\end{align*}
 We further assume that $\mathbf{U}$ and $\mathbf{v}$ are deterministic constants. Consider the variance of the output under this distribution:
 \begin{align}
     \mathrm{Var}[y_k (\mathbf{x})] &= \sum_i \mathrm{Var}[W_{ki}] \phi(a_i)^2 + \mathrm{Var}[b_k]. \label{sum_variance}
 \end{align}
 Equation (\ref{sum_variance}) is justified since each weight is independent under $q$. This variance is a measure of the uncertainty in the output at $\mathbf{x}$ represented by the approximate posterior $q$. Consider the Hessian of this variance $\mathbf{H}$, where $H_{nm} = \partial_{x_n}\partial_{x_m} \mathrm{Var}[y_k(\mathbf{x})]$. Taking derivatives, we have:
 \begin{align*}
     H_{nm} &=  \sum_i 2 \mathrm{Var}[W_{ki}] \big(\phi(a_i)\phi''(a_i) + \phi'(a_i)^2 \big) U_{in}U_{im}\\
     \mathbf{H} &= \sum_i 2\mathrm{Var}[W_{ki}] \big(\phi(a_i)\phi''(a_i) + \phi'(a_i)^2 \big) \mathbf{u}_i \mathbf{u}^{\mathsf{T}}_i,
 \end{align*}
 where $\mathbf{u}_i$ is the column vector whose elements are the $i$th row of $\mathbf{U}$. Since $\mathbf{H}$ is a sum of outer products, it will be positive semi-definite (PSD) if $\phi(a_i)\phi''(a_i) + \phi'(a_i)^2 \geq 0$ for all $i$. This is the case for ReLU nonlinearities. The first and second derivatives of the ReLU do not exist at $a_i = 0$. However if we consider $\phi''$ to be a bump function of arbitrarily small width and area 1, then all these derivatives exist and $\phi(a_i)\phi''(a_i)$ is non-negative. Since the Hessian of $\mathrm{Var}[y_k(\mathbf{x})]$ is PSD, it follows that $\mathrm{Var}[y_k(\mathbf{x})]$ is a convex function of $\mathbf{x}$.\footnote{This argument can be made rigorous by constructing a sequence of networks with non-linearities $\phi_n$ such that $\phi_n''$ is a triangular function at zero with area 1 and width $\epsilon/n$. Each network will have a convex output variance, and the variance of these networks converges pointwise to the variance of a ReLU network. Since a pointwise limit of convex functions is convex, the result holds for ReLU networks.} Therefore it is impossible for this kind of posterior to exhibit greater uncertainty in between two regions of low uncertainty. 
 
\begin{figure}
\vskip 0.2in
\begin{center}
\centerline{\includegraphics[width=0.9\columnwidth]{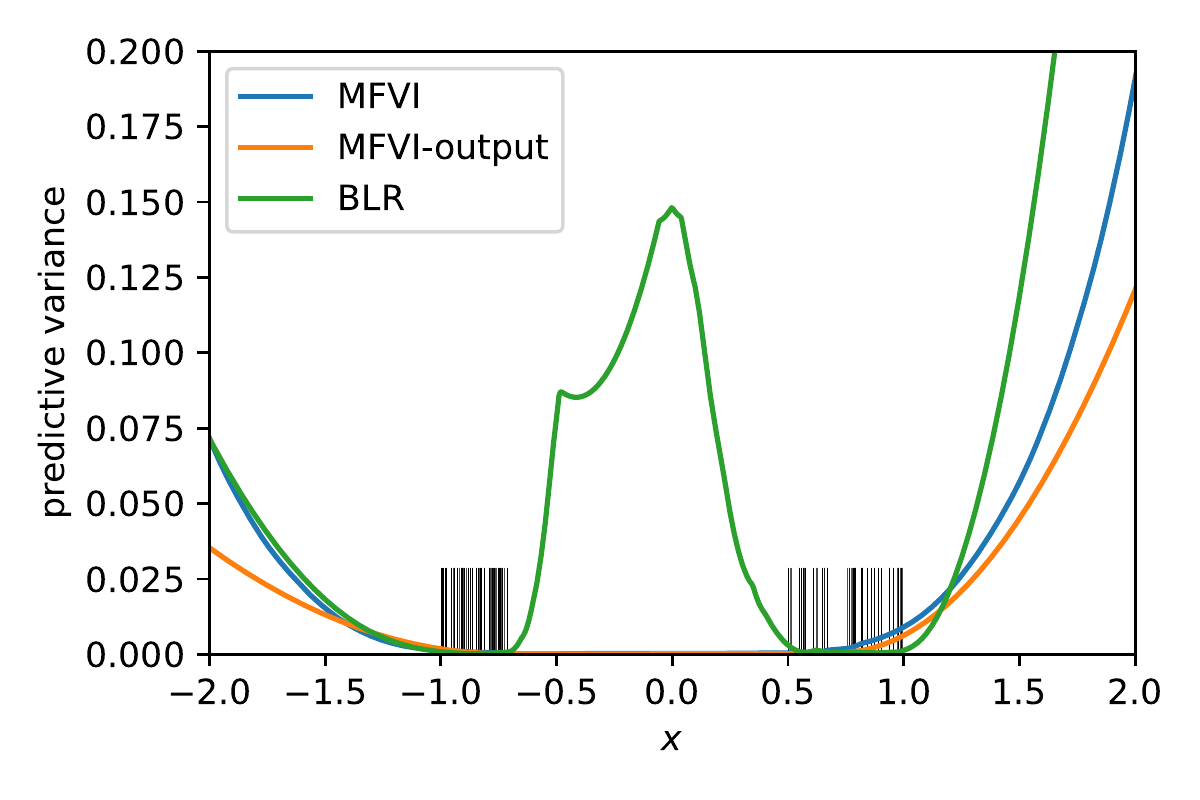}}
\caption{Predictive variances (without observation noise) on the 1D dataset. Black lines show $x$-locations of the data. }
\label{variance_comparison}
\end{center}
\vskip -0.4in
\end{figure}

 To investigate the relevance of this result to the standard case where the input parameters $(\mathbf{U}, \mathbf{v})$ are not deterministic but are also mean-field, we train three ReLU BNNs on the 1D dataset in Figure \ref{tanh_comparison}: (i) mean-field VI on all parameters (MVFI), (ii) maximum-likelihood on the input parameters and mean-field on the output parameters (MFVI-output) and (iii) maximum-likelihood on the input parameters followed by Bayesian linear regression on the output parameters (BLR). Results are shown in Figure \ref{variance_comparison}. MFVI-output has convex variance, as predicted. MFVI also has convex variance, even though its input parameters are mean-field. BLR with its full-covariance Gaussian posterior shows increased in-between uncertainty even though its input weights are deterministic, showing that it is the mean-field assumption on the output parameters that is responsible for severely restricting the expressiveness of the predictive uncertainty. Further work is required to characterise the expressiveness of mean field distributions with deeper networks.
 
\section{Analysis of Uncertainty in Toy Case}
\label{toy_explanation}
To gain intuition for why MFVI fails to provide in-between uncertainty, we consider a toy example involving a single hidden layer network with two ReLU hidden units mapping $x \in \mathbb{R} \rightarrow y \in \mathbb{R}$:
\begin{align*}
    y(x) = W_1 \phi (U_1 x + v_1) + W_2 \phi (U_2 x + v_2) + b.
\end{align*}
Here $W_1, W_2$ and $b$ are the output weights and bias, and $U_1, U_2$ and $v_1, v_2$ are the input weights and biases. Consider the case where $W_1, W_2, U_1, U_2$ are all deterministic and positive so that $y(x)$ is non-decreasing. Then:
\begin{align*}
    y(x)=
    \begin{cases}
    b & \mathrm{(I)}\\
    W_1U_1x + W_1v_1 + b & \mathrm{(II)}\\
    (W_1U_1 + W_2U_2)x + W_1v_1 + W_2v_2 + b & \mathrm{(III)}
    \end{cases}
\end{align*}
where $x< -\frac{v_1}{U_1}$ in region (I), $-\frac{v_1}{U_1} \leq x < -\frac{v_2}{U_2}$ in region (II) and $x\geq -\frac{v_2}{U_2}$ in region (III). Consider a simple observed dataset with many points at $(x_1, y_1)$ and $(x_2, y_2)$ where $x_2 > x_1$ and $y_2 > y_1$. A reasonable Bayesian posterior predictive would have low uncertainty around $x_1$ and $x_2$, but large uncertainty in between. To first fit this dataset with deterministic weights, we could place $x_1$ in region (I) and $x_2$ in region (III). Then to fit the $y$-values we must set
\begin{align}
    b &\approx y_1, \\
    (W_1U_1 + W_2U_2)x_2 &+ W_1v_1 + W_2v_2 + b \approx y_2. \label{fit_equation}
\end{align}

\begin{figure}
\vskip 0.in
\begin{center}
\centerline{\includegraphics[width=0.8\columnwidth]{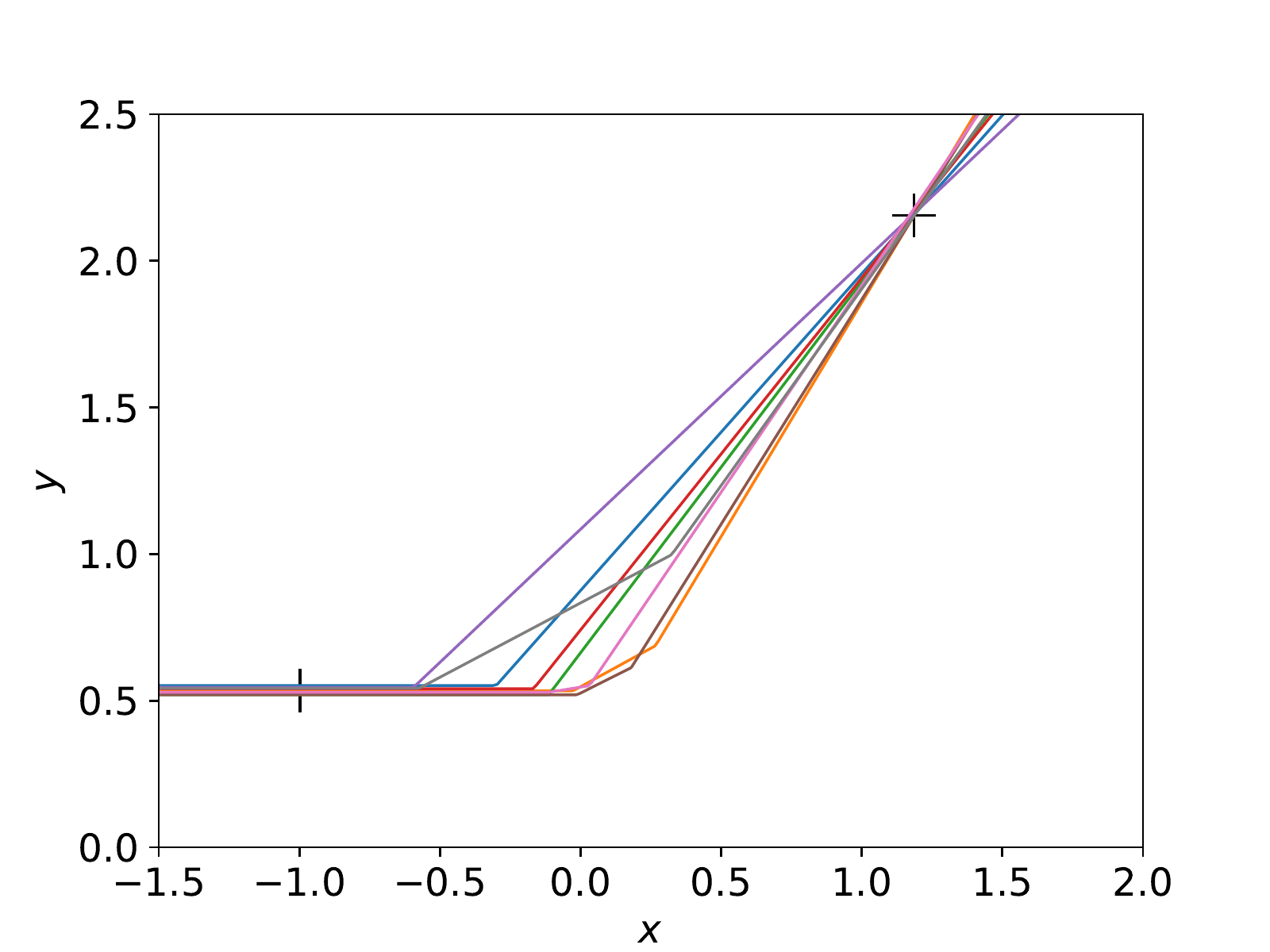}}
\caption{Samples from a 2-hidden unit neural network obtained by HMC. Notice how the position of the kinks varies between samples, leading to larger uncertainty in between the 2 datapoints $(x_1,y_1)$ and $(x_2,y_2)$, marked by black crosses. (For some of these samples, only one kink is between $x_1$ and $x_2$; the other is to the left of $x_1$.)}
\label{2_points}
\end{center}
\vskip -0.3in
\end{figure}

There are many settings of $W_1, U_1, W_2, U_2, v_1$ and $v_2$ that satisfy Equation \ref{fit_equation}. Consider choosing one such setting as a point estimate. To obtain a Bayesian method, we would now like to increase our uncertainty in the parameters. In particular, we should have relatively large uncertainty in the position of the `kinks' $-\frac{v_1}{U_1}$ and $-\frac{v_2}{U_2}$ since they can take any values between $x_1$ and $x_2$ and fit the data equally well.\footnote{Here we assume a reasonably broad prior such that the prior probabilities of the kink locations are roughly uniform over the range $[x_1, x_2]$.} This corresponds to having large uncertainty between two regions of low uncertainty ($x_1$ and $x_2$), as in Figure \ref{2_points}. To express this, we could relax the distribution over, say, $v_2$ from a delta function to a Gaussian with positive variance. However, injecting randomness in $v_2$ jeopardises the fit in region (III) since $v_2$ is involved in Equation \ref{fit_equation}. The only way to express predictive uncertainty between $x_1$ and $x_2$ and \emph{still} fit the data is to have the values of $W_1, U_1, W_2, U_2$ and $v_1$ \emph{compensate} for any change in $v_2$ such that Equation \ref{fit_equation} still holds. In other words, we need strong dependencies between the parameters to simultaneously fit the data regions and express predictive uncertainty in the in-between region.\footnote{The in-between uncertainty seen in Figure 3 in \cite{duvenaud2015black} is seemingly an exception. However in that case radial basis function non-linearities were used. Since these have only local effects, the argument here does not apply.} 

The mean-field approximation assumes that there are no dependencies. Therefore any parameter randomness used to express increased in-between uncertainty leads to uncontrolled variations of $y(x)$ in the data region. Hence Equation \ref{fit_equation} is not satisfied. There are two possibilities: either the data fit will be poor or the variances will be minimised (leading to a large penalty in the KL term in the ELBO). In practice MFVI finds a solution that prunes out hidden units, allowing it to fit the data with the minimum number of variances set to zero \cite{trippe2018overpruning}.

\section{Extra Results and Experimental Details}
\label{appendix_extra}
\begin{figure*}
\vskip 0.2in
\begin{center}
\centerline{\includegraphics[width=2\columnwidth]{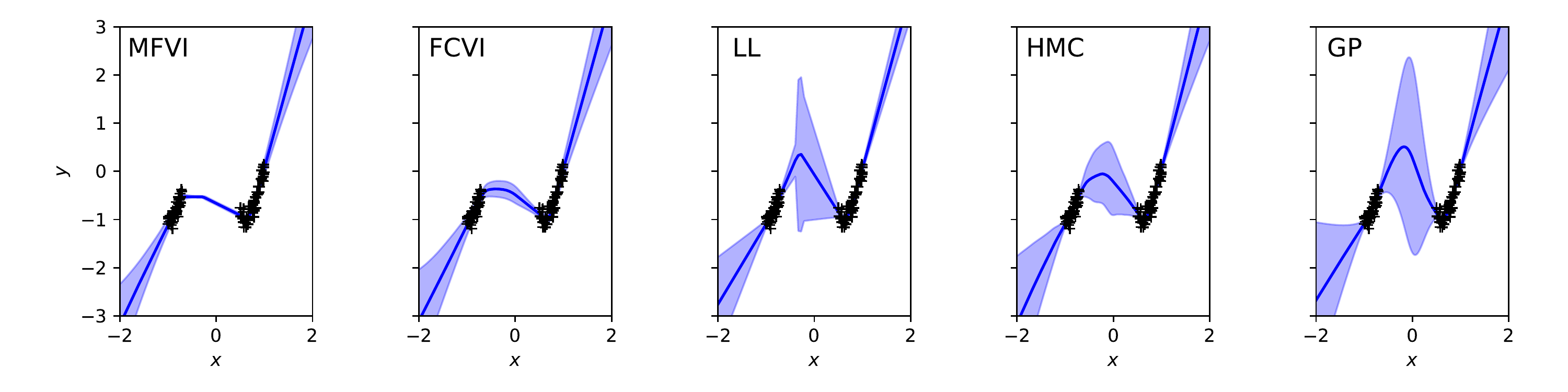}}
\caption{Mean and two standard deviation bars of the predictive distribution for $f_{\theta}(x)$ (without output noise) using ReLU activations. }
\label{relu_comparison}
\end{center}
\vskip -0.2in
\end{figure*}

\textbf{Synthetic 1D Dataset}. We use single hidden layer BNNs with 50 hidden units. We include results for ReLU activations in Figure \ref{relu_comparison}. To verify that MFVI's lack of in-between uncertainty is due to approximate inference and not the model class, we include a Gaussian Process (GP) using the kernel for a BNN with infinitely many ReLU hidden units \cite{lee2017deep,cho2009kernel}. Note LL shows strange discontinuous behaviour in its uncertainty. This is because the non-smooth ReLU function makes the gradient $g(\mathbf{x})$ in equation (\ref{linearisation}) discontinuous. Similar behaviour is seen in \cite{snoek2015scalable}.

We use independent $\mathcal{N}(0,1)$ priors on the biases and $\mathcal{N}(0,\frac{\omega^2}{H})$ priors on the weights; $H$ is the number of inputs to the weight matrix. This scaling is chosen so that the GP limit exists \cite{neal2012bayesian}. $\omega$ is set to $4$. To optimise Laplace, MFVI and FCVI we use ADAM \cite{kingma2014adam} with learning rate 0.001 and 20,000 epochs. We use the entire dataset for each batch. For MFVI, weight means were initialised from $\mathcal{N}(0,1/\sqrt{4n_{outputs}})$ and all variances were initialised to $10^{-5}$. Bias means were initialised to zero. The local reparameterisation trick \cite{kingma2015variational} was used. For FCVI, the Cholesky decomposition of the covariance matrix was parameterised as a lower triangular matrix, with the diagonal entries made positive by exponentiating them. The diagonal entries were initialised to $\log(0.05)$ and the off-diagonals were initialised to 0. The mean vector was initialised from $\mathcal{N}(0, 0.1)$. For both MFVI and FCVI we approximate the ELBO during training with 32 samples. For HMC, the number of leapfrog steps was chosen uniformly between 5 and 10, and the step size uniformly sampled from $[0.001, 0.0015]$. The chain was burned in for 10,000 iterations and samples were collected during the next 20,000 iterations. For MFVI, FCVI and HMC, the error bars in Figures \ref{tanh_comparison} and \ref{relu_comparison} were estimated with 100 samples.

\textbf{UCI Datasets}. All BNNs had 50 neurons per hidden layer. Inputs and outputs were normalised to zero mean and unit variance. Hyperparameters were optimised by grid search on a validation set that consisted of $10\%$ of the training set. The best hyperparameters were used to train again on the training set  with  validation  set  combined. This was repeated for each split. Minibatches were randomly selected from the training set with replacement. For MAP and Laplace, all parameters had independent $\mathcal{N}(0,\omega^2)$ priors. For Laplace, minibatch size was $100$. The hyperparameters optimised were: $\omega$: $[1, 2]$, learning rate: $[0.01, 0.001]$, number of epochs: $[40, 100, 200, 400]$. For MAP, the same ranges were searched, except the number of epochs was $[20, 40, 100]$, since we expected MAP to favour early stopping. For both methods, $\log(\sigma_o^2)$ was initialised to $-1$ and learned by maximum likelihood. For FCVI and MFVI, we used independent $\mathcal{N}(0,1)$ priors on all parameters. The hyperparameters searched for MFVI were: minibatch size: $[32, 100]$, learning rate: $[0.01, 0.001]$, number of epochs: $[500, 1000, 2000]$ for smaller datasets (boston, concrete, energy, wine, yacht) and $[50, 100, 200]$ for larger ones (kin8nm, naval, power, protein). The same ranges were used for FCVI except the learning rate was fixed to $0.001$. The ELBO was approximated with $32$ samples. For MFVI, weight means were initialised from $\mathcal{N}(0,1/\sqrt{4n_{out}})$ and all variances initialised to $10^{-5}$. Bias means were initialised to zero. For FCVI, the mean vector and covariance matrix of all the parameters in the network were optimised to maximise the ELBO. The Cholesky decomposition of the covariance matrix was parameterised directly as a lower triangular matrix, with the diagonal entries constrained to be positive by exponentiating them. The diagonal entries were initialised to $\log(10^{-5})$ and the off-diagonal entries were initialised to $0$. The mean vector was initialised randomly from $\mathcal{N}(0, 0.1)$. For MFVI, FCVI and sampled Laplace, test log-likelihoods were computed by sampling $100$ times from the approximate posterior. $\log(\sigma_o^2)$ was initialised to $-1$ and learned by optimising the ELBO. 

\begin{figure}
\begin{center}
\centerline{\includegraphics[width=0.9\columnwidth]{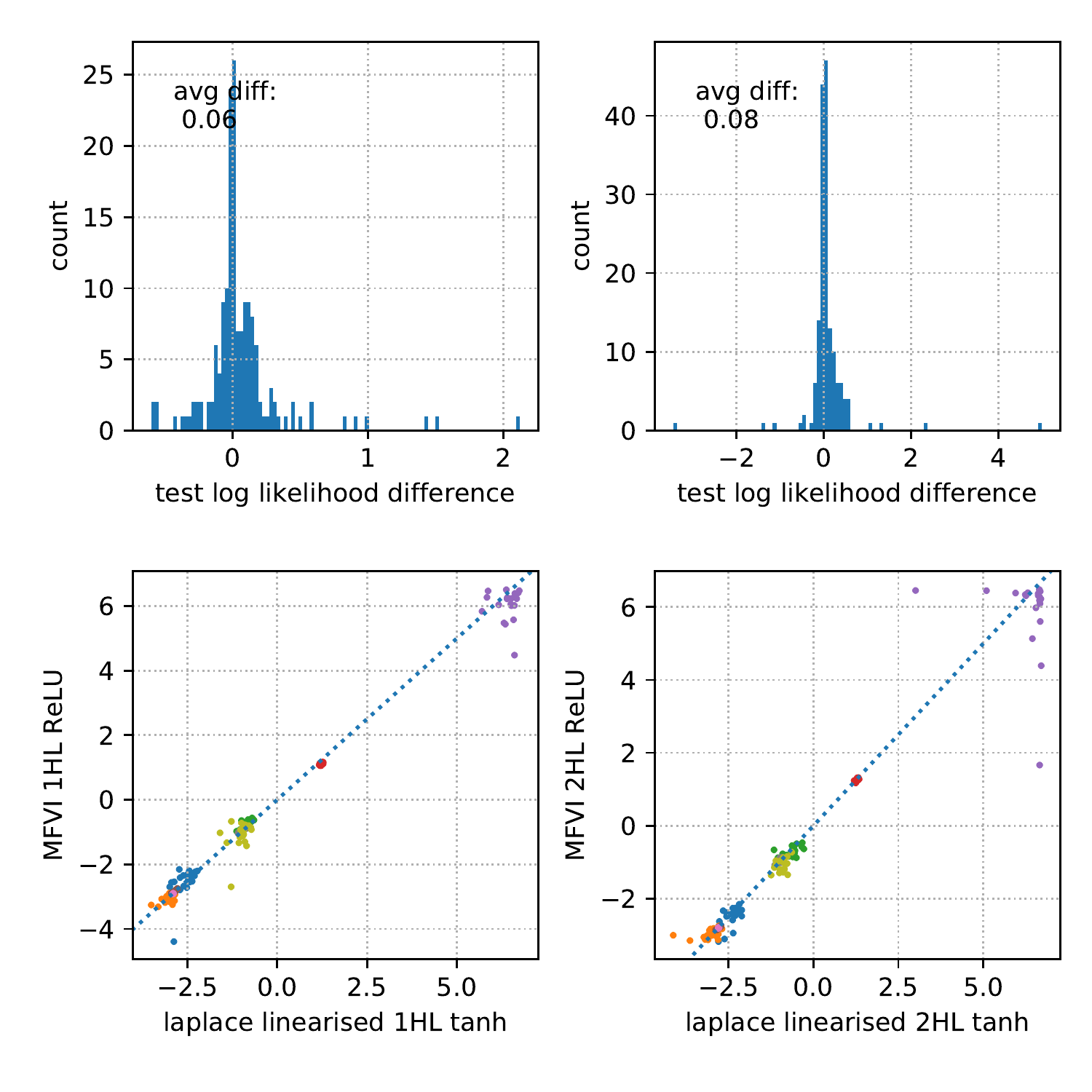}}
\caption{Comparison of MFVI-ReLU and linearised Laplace tanh on the standard splits. Positive difference means Laplace performs better than MFVI.}
\label{MFVIvslap_standard}
\end{center}
\vskip -0.2in
\end{figure}

\begin{figure}
\begin{center}
\centerline{\includegraphics[width=0.9\columnwidth]{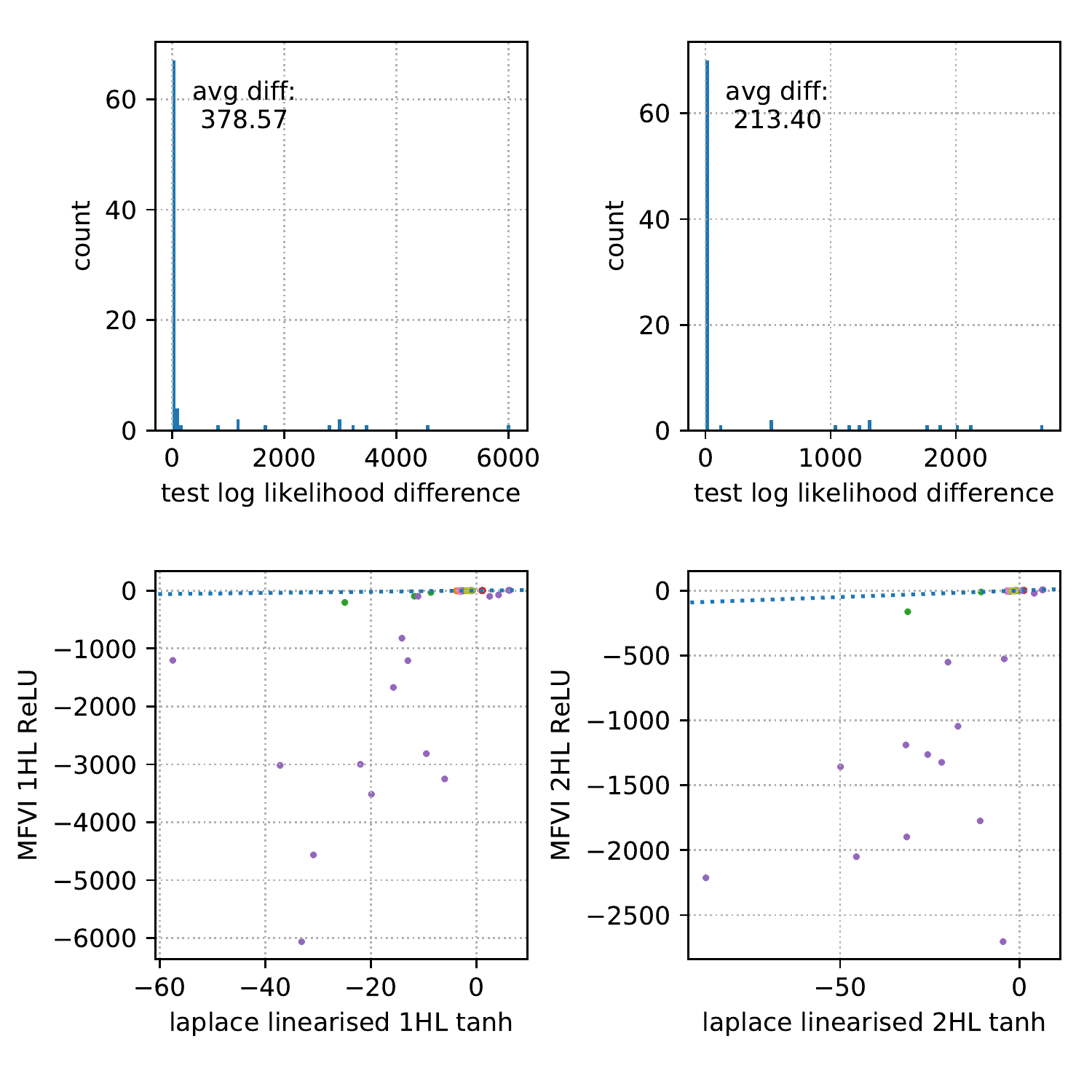}}
\caption{Comparison of MFVI-ReLU and linearised Laplace tanh on the gap splits. Positive difference means Laplace performs better than MFVI. MFVI fails catastrophically on energy and naval.}
\label{MFVIvslap_gap}
\end{center}
\vskip -0.2in
\end{figure}

\begin{figure}
\begin{center}
\centerline{\includegraphics[width=0.9\columnwidth]{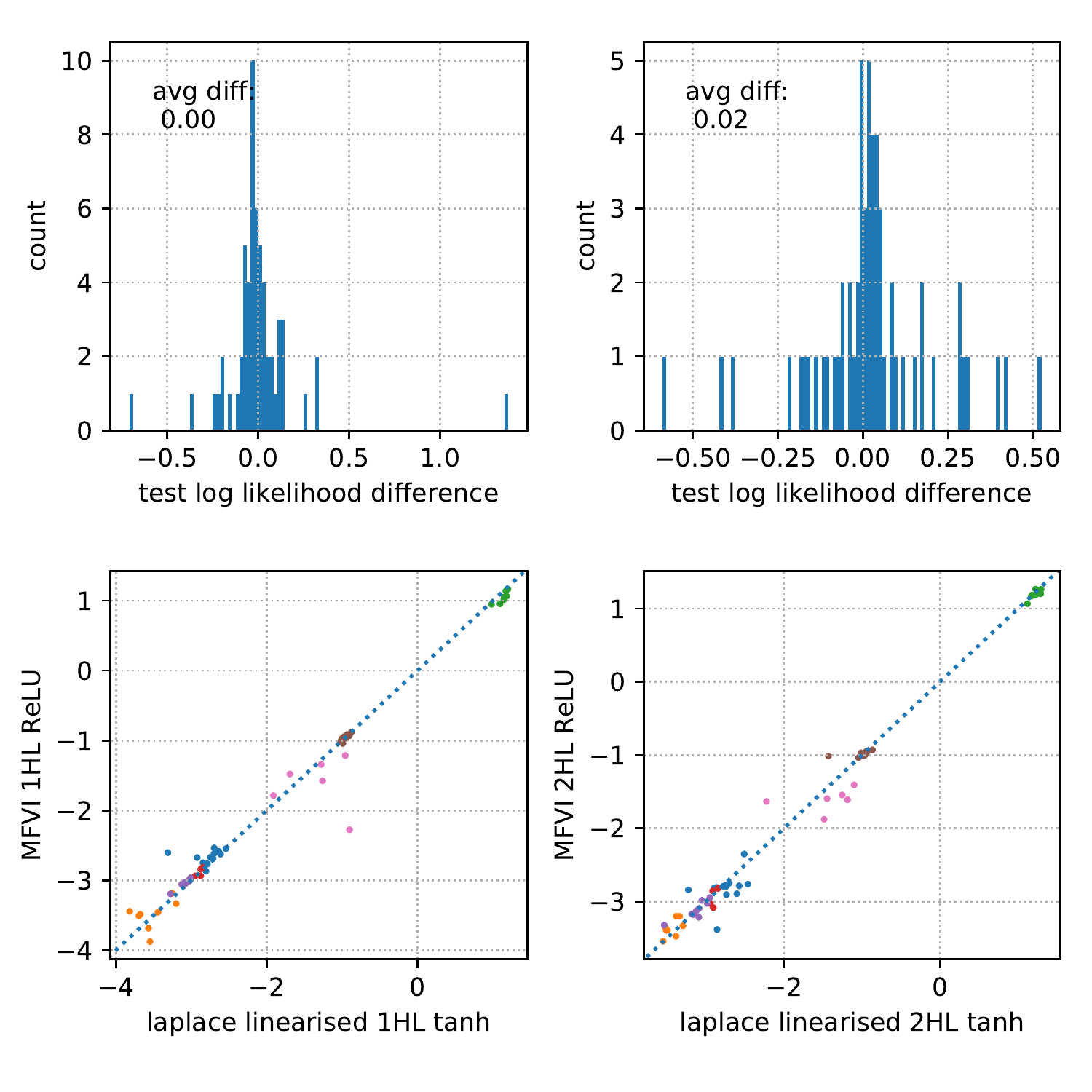}}
\caption{Same as Figure \ref{MFVIvslap_gap} but with energy and naval removed. Positive difference means Laplace performs better than MFVI. The two methods now perform comparably.}
\label{MFVIvslap_gap_removed}
\end{center}
\vskip -0.2in
\end{figure}

Full results are given in Tables \ref{standard_split_table} and \ref{gap_split_table}. We also provide pairwise comparisons of LL versus MFVI-ReLU on the standard splits and the gap splits in Figures \ref{MFVIvslap_standard}, \ref{MFVIvslap_gap} \& \ref{MFVIvslap_gap_removed}. Each point corresponds to one test log-likelihood. Each colour represents a different dataset. The histogram shows the log likelihood of the method on the $x$-axis minus that of the method on the $y$-axis. The dotted blue line is the line $y=x$.

\begin{table*}[t]
\caption{Average test log likelihoods for standard splits. Best results within one standard error in bold.}
\label{standard_split_table}
\vskip 0.15in
\begin{center}
\begin{small}
\begin{sc}
\resizebox{\textwidth}{!}{%
\begin{tabular}{ l c c c c c c c c c c r  }
\toprule
   model/dataset& boston& concrete& energy& kin8nm&  naval& power& protein& wine& yacht\\
\midrule
MAP 1HL relu& $-2.71 \pm 0.10$& $-3.17 \pm 0.04$& $-1.02 \pm 0.05$& $1.09 \pm 0.01$& $5.75 \pm 0.05$& $-2.82 \pm 0.01$& $-2.92 \pm 0.00$& $-0.98 \pm 0.02$& $-2.30 \pm 0.02$\\
MAP 1HL tanh& $-2.69 \pm 0.09$& $-3.22 \pm 0.03$& $-1.12 \pm 0.05$& $1.15 \pm 0.01$& $5.48 \pm 0.05$& $-2.85 \pm 0.01$& $-2.92 \pm 0.00$& $-0.99 \pm 0.02$& $-2.51 \pm 0.02$\\
MAP 2HL relu& $-2.67 \pm 0.11$& $-3.23 \pm 0.05$& $-0.92 \pm 0.05$& $1.16 \pm 0.01$& $6.18 \pm 0.05$& $\mathbf{-2.79 \pm 0.01}$& $-2.83 \pm 0.00$& $-1.00 \pm 0.02$& $-2.31 \pm 0.01$\\
MAP 2HL tanh& $-2.80 \pm 0.12$& $-3.33 \pm 0.05$& $-0.86 \pm 0.02$& $1.24 \pm 0.01$& $5.72 \pm 0.10$& $-2.81 \pm 0.01$& $-2.83 \pm 0.01$& $-0.99 \pm 0.02$& $-2.22 \pm 0.01$\\
MFVI 1HL relu& $-2.55 \pm 0.10$& $-3.08 \pm 0.03$& $-0.75 \pm 0.03$& $1.12 \pm 0.01$& $6.04 \pm 0.11$& $-2.82 \pm 0.01$& $-2.91 \pm 0.00$& $\mathbf{-0.96 \pm 0.01}$& $-1.09 \pm 0.10$\\
MFVI 1HL tanh& $-2.61 \pm 0.05$& $-3.12 \pm 0.02$& $-0.82 \pm 0.03$& $1.10 \pm 0.01$& $5.79 \pm 0.14$& $-2.86 \pm 0.01$& $-2.93 \pm 0.00$& $\mathbf{-0.97 \pm 0.01}$& $-1.32 \pm 0.05$\\
MFVI 2HL relu& $\mathbf{-2.49 \pm 0.06}$& $ \mathbf{ -2.97 \pm 0.03 }$& $\mathbf{-0.72 \pm 0.03}$& $1.26 \pm 0.01$& $5.88 \pm 0.24$& $\mathbf{-2.80 \pm 0.01}$& $-2.81 \pm 0.00$& $ \mathbf{ -0.95 \pm 0.01 }$& $-1.08 \pm 0.04$\\
MFVI 2HL tanh& $-2.70 \pm 0.06$& $-3.20 \pm 0.01$& $-1.60 \pm 0.07$& $1.19 \pm 0.01$& $6.08 \pm 0.05$& $-2.87 \pm 0.01$& $-2.88 \pm 0.01$& $-0.98 \pm 0.01$& $-1.87 \pm 0.03$\\
FCVI 1HL relu& $-2.58 \pm 0.03$& $-3.17 \pm 0.01$& $-1.42 \pm 0.01$& $0.87 \pm 0.00$& $3.82 \pm 0.02$& $-2.86 \pm 0.01$& $-3.16 \pm 0.04$& $\mathbf{-0.96 \pm 0.01}$& $-1.67 \pm 0.01$\\
FCVI 1HL tanh& $-2.77 \pm 0.02$& $-3.31 \pm 0.01$& $-1.76 \pm 0.02$& $0.83 \pm 0.00$& $3.37 \pm 0.00$& $-2.88 \pm 0.01$& $-3.02 \pm 0.00$& $-0.99 \pm 0.01$& $-2.00 \pm 0.01$\\
LL 1HL tanh& $-2.57 \pm 0.05$& $-3.05 \pm 0.04$& $-0.82 \pm 0.03$& $1.23 \pm 0.01$& $ \mathbf{ 6.40 \pm 0.06 }$& $-2.83 \pm 0.01$& $-2.89 \pm 0.00$& $\mathbf{-0.97 \pm 0.02}$& $-1.01 \pm 0.05$\\
SL 1HL tanh& $-2.85 \pm 0.05$& $-3.27 \pm 0.01$& $-2.77 \pm 0.02$& $1.15 \pm 0.01$& $3.19 \pm 0.05$& $-3.41 \pm 0.01$& $-2.90 \pm 0.00$& $-1.01 \pm 0.01$& $-3.75 \pm 0.02$\\
LL 2HL tanh& $ \mathbf{ -2.40 \pm 0.04 }$& $\mathbf{-3.07 \pm 0.07}$& $ \mathbf{ -0.65 \pm 0.05 }$& $ \mathbf{ 1.29 \pm 0.01 }$& $\mathbf{6.29 \pm 0.19}$& $ \mathbf{-2.79 \pm 0.01 }$& $\mathbf{-2.79 \pm 0.00}$& $-0.98 \pm 0.01$& $ \mathbf{ -0.92 \pm 0.03 }$\\
SL 2HL tanh& $-2.82 \pm 0.03$& $-3.29 \pm 0.02$& $-2.67 \pm 0.02$& $1.26 \pm 0.01$& $3.80 \pm 0.02$& $-3.23 \pm 0.01$& $ \mathbf{ -2.79 \pm0.00 }$& $-1.07 \pm 0.05$& $-3.63 \pm 0.04$\\
\bottomrule
\end{tabular}}
\end{sc}
\end{small}
\end{center}
\end{table*}

\begin{table*}[t]
\caption{Average test log likelihoods for gap splits. Best results within one standard error in bold.}
\label{gap_split_table}
\begin{center}
\begin{small}
\begin{sc}
\resizebox{\textwidth}{!}{%
\begin{tabular}{ l c c c c c c c c c c r  }
\toprule
   model/dataset& boston& concrete& energy& kin8nm&  naval& power& protein& wine& yacht\\
\midrule
MAP 1HL relu& $-2.88 \pm 0.09$& $-3.62 \pm 0.07$& $-67.81 \pm 39.06$& $1.02 \pm 0.03$& $-885.51 \pm 187.94$& $ \mathbf{ -2.84 \pm 0.02 }$& $\mathbf{-3.05 \pm 0.02}$& $\mathbf{-0.98 \pm 0.02}$& $-2.37 \pm 0.03$\\
MAP 1HL tanh& $-2.90 \pm 0.06$& $-3.64 \pm 0.05$& $-104.53 \pm 61.60$& $1.09 \pm 0.03$& $-260.25 \pm 47.20$& $-2.89 \pm 0.02$& $\mathbf{-3.08 \pm 0.02}$& $\mathbf{-0.97 \pm 0.01}$& $-2.63 \pm 0.21$\\
MAP 2HL relu& $-2.83 \pm 0.09$& $-3.47 \pm 0.04$& $-37.51 \pm 22.79$& $1.11 \pm 0.03$& $-742.49 \pm 168.32$& $\mathbf{-2.86 \pm 0.04}$& $-3.10 \pm 0.03$& $\mathbf{-0.99 \pm 0.02}$& $-2.41 \pm 0.04$\\
MAP 2HL tanh& $-2.91 \pm 0.06$& $-3.68 \pm 0.17$& $-113.12 \pm 66.11$& $1.13 \pm 0.05$& $-219.61 \pm 40.19$& $-2.90 \pm 0.02$& $\mathbf{-3.08 \pm 0.02}$& $\mathbf{-0.97 \pm 0.02}$& $-2.27 \pm 0.03$\\
MFVI 1HL relu& $ \mathbf{ -2.69 \pm 0.03 }$& $-3.50 \pm 0.07$& $-42.43 \pm 24.25$& $1.05 \pm 0.03$& $-1961.52 \pm 447.82$& $\mathbf{-2.88 \pm 0.03}$& $ \mathbf{ -3.04 \pm 0.02 }$& $ \mathbf{ -0.96 \pm 0.01 }$& $\mathbf{-1.61 \pm 0.14}$\\
MFVI 1HL tanh& $-2.88 \pm 0.10$& $-3.64 \pm 0.10$& $-100.16 \pm 59.45$& $1.01 \pm 0.02$& $-385.33 \pm 82.47$& $\mathbf{-2.88 \pm 0.02}$& $\mathbf{-3.08 \pm 0.02}$& $\mathbf{-0.96 \pm 0.01}$& $-1.77 \pm 0.07$\\
MFVI 2HL relu& $-2.82 \pm 0.06$& $ \mathbf{ -3.36 \pm 0.04 }$& $\mathbf{-23.42 \pm 18.71}$& $\mathbf{1.20 \pm 0.02}$& $-1119.13 \pm 211.08$& $-2.95 \pm 0.06$& $-3.12 \pm 0.04$& $\mathbf{-0.98 \pm 0.01}$& $-1.61 \pm 0.06$\\
MFVI 2HL tanh& $-2.92 \pm 0.08$& $-3.60 \pm 0.03$& $\mathbf{-29.98 \pm 23.40}$& $1.10 \pm 0.03$& $-205.93 \pm 43.89$& $\mathbf{-2.88 \pm 0.02}$& $\mathbf{-3.07 \pm 0.02}$& $-1.01 \pm 0.01$& $-1.87 \pm 0.13$\\
FCVI 1HL relu& $\mathbf{-2.72 \pm 0.03}$& $\mathbf{-3.38 \pm 0.02}$& $\mathbf{-16.35 \pm 8.32}$& $0.83 \pm 0.01$& $-2.01 \pm 1.09$& $-2.89 \pm 0.01$& $-3.09 \pm 0.01$& $\mathbf{-0.96 \pm 0.01}$& $-2.06 \pm 0.08$\\
FCVI 1HL tanh& $-2.92 \pm 0.04$& $-3.48 \pm 0.04$& $\mathbf{-9.22 \pm 4.18}$& $0.78 \pm 0.01$& $\mathbf{2.18 \pm 0.13}$& $-2.88 \pm 0.01$& $\mathbf{-3.07 \pm 0.01}$& $-1.01 \pm 0.01$& $-2.46 \pm 0.13$\\
LL 1HL tanh& $-2.79 \pm 0.05$& $-3.53 \pm 0.07$& $\mathbf{-6.49 \pm 2.81}$& $1.14 \pm 0.02$& $-15.66 \pm 4.26$& $-2.89 \pm 0.02$& $\mathbf{-3.09 \pm 0.03}$& $\mathbf{-0.96 \pm 0.01}$& $ \mathbf{ -1.33 \pm 0.15 }$\\
SL 1HL tanh& $-3.03 \pm 0.03$& $-3.67 \pm 0.07$& $\mathbf{-14.08 \pm 7.03}$& $1.06 \pm 0.02$& $ \mathbf{ 2.20 \pm 0.14 }$& $-3.45 \pm 0.03$& $\mathbf{-3.08 \pm 0.02}$& $-1.00 \pm 0.01$& $-3.95 \pm 0.17$\\
LL 2HL tanh& $\mathbf{-2.75 \pm 0.05}$& $\mathbf{-3.43 \pm 0.03}$& $ \mathbf{ -6.34 \pm 3.49 }$& $ \mathbf{ 1.22 \pm 0.02 }$& $-20.76 \pm 6.05$& $-2.90 \pm 0.02$& $-3.13 \pm 0.05$& $\mathbf{-1.01 \pm 0.04}$& $\mathbf{-1.45 \pm 0.15}$\\
SL 2HL tanh& $-3.01 \pm 0.05$& $-3.62 \pm 0.04$& $\mathbf{-17.17 \pm 8.88}$& $\mathbf{1.20 \pm 0.04}$& $-51.99 \pm 13.75$& $-3.36 \pm 0.04$& $\mathbf{-3.11 \pm 0.05}$& $-1.12 \pm 0.08$& $-3.61 \pm 0.06$\\
\bottomrule
\end{tabular}}
\end{sc}
\end{small}
\end{center}
\end{table*}

\end{document}